\documentclass[10pt,twocolumn,letterpaper]{article}

\usepackage{cvpr}
\usepackage{times}
\usepackage{epsfig}
\usepackage{graphicx}
\usepackage{amsmath}
\usepackage{amssymb}
\usepackage{multirow}
\usepackage{booktabs}
\usepackage{makecell}

\usepackage[pagebackref=true,breaklinks=true,letterpaper=true,colorlinks,bookmarks=false]{hyperref}

\cvprfinalcopy 

\def\cvprPaperID{****} 

\ifcvprfinal\fi
\begin{document}
\title{SG-RIFE: Semantic-Guided Real-Time Intermediate Flow Estimation\\with Diffusion-Competitive Perceptual Quality}

\author{Pan Ben Wong, Chengli Wu,  Hanyue Lu\\
Georgia Institute of Technology\\
{\tt\small \{pwong48, cwu602, hlu350\}@gatech.edu}\\
}

\maketitle

\begin{abstract}
Real-time Video Frame Interpolation (VFI) has long been dominated by flow-based methods like RIFE, which offer high throughput but often fail in complicated scenarios involving large motion and occlusion. Conversely, recent diffusion-based approaches (\eg Consec. BB) achieve state-of-the-art perceptual quality but suffer from prohibitive latency, rendering them impractical for real-time applications. To bridge this gap, we propose \textbf{Semantic-Guided RIFE (SG-RIFE)}. Instead of training from scratch, we introduce a parameter-efficient fine-tuning strategy that augments a pre-trained RIFE backbone with semantic priors from a frozen DINOv3 Vision Transformer. We propose a \textbf{Split-Fidelity Aware Projection Module (Split-FAPM)} to compress and refine high-dimensional features, and a \textbf{Deformable Semantic Fusion (DSF)} module to align these semantic priors with pixel-level motion fields. Experiments on SNU-FILM demonstrate that semantic injection provides a decisive boost in perceptual fidelity. SG-RIFE outperforms diffusion-based LDMVFI in FID/LPIPS and achieves quality comparable to Consec. BB on complex benchmarks while running significantly faster, proving that semantic consistency enables flow-based methods to achieve diffusion-competitive perceptual quality in near real-time.
\end{abstract}

\section{Introduction}
\label{sec:intro}
\begin{figure}[t]
  \centering
  \includegraphics[width=\linewidth]{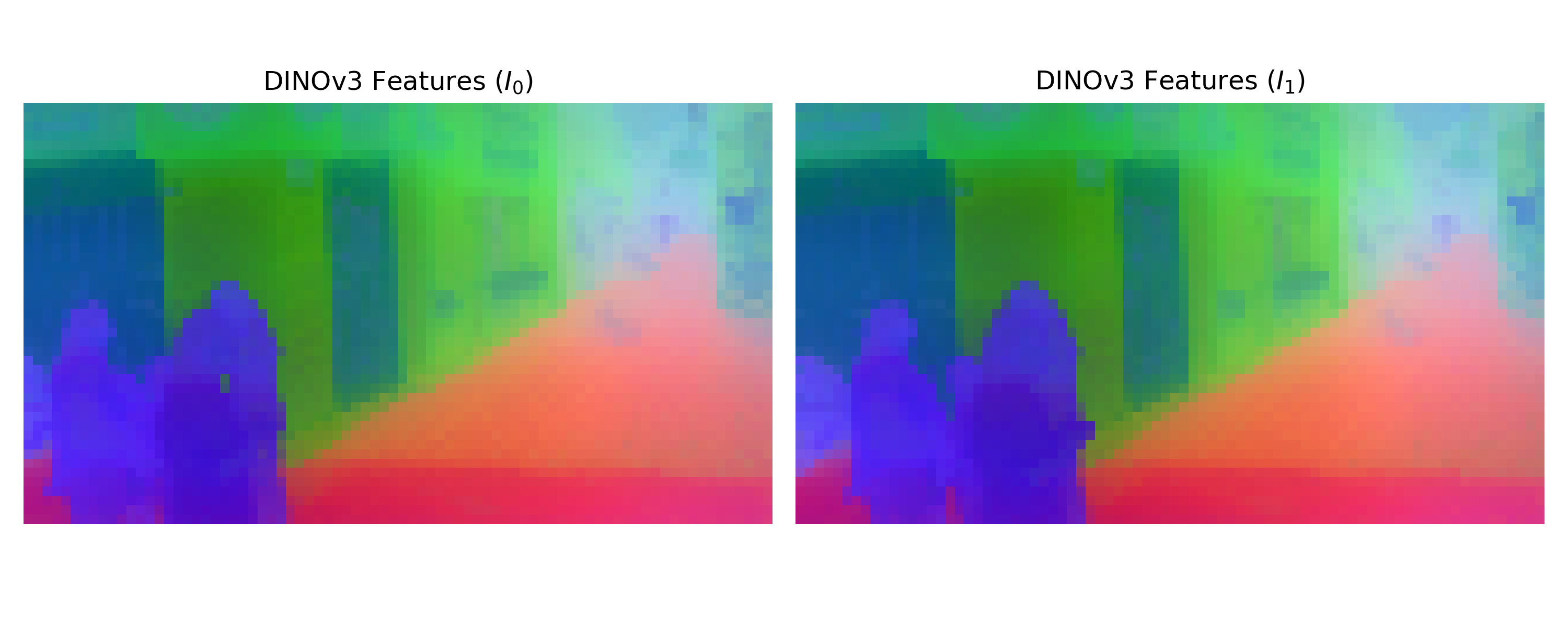} 
  \caption{\textbf{Semantic Stability Analysis.} DINOv3 features for two consecutive frames using a shared PCA basis. Despite large displacement between $I_0$ (Left) and $I_1$ (Right), the semantic representations of moving objects (\eg, the pedestrians) have consistent color signatures. This stability allows SG-RIFE to maintain object identity where traditional flow-based context typically degrades.}
  \label{fig:motivation}
\end{figure}

Video Frame Interpolation (VFI) synthesizes intermediate frames between consecutive sequences and is critical for applications like slow-motion playback and video streaming. Currently, the field faces a stark trade-off: efficient flow-based methods (\eg RIFE~\cite{huang2020rife}) treat images as grids of colored pixels and often produce blur or ghosting in complex motion, while generative diffusion methods (\eg Consec. BB~\cite{BBD}) produce sharp details but require seconds to generate a single frame.

We hypothesize that the limitations of real-time models stem not from a lack of temporal context, but from a lack of semantic understanding. As shown in Figure~\ref{fig:motivation}, semantic representations of moving objects have consistent signatures even across large displacements. By injecting high-level object priors, a model should be able to inpaint plausible textures even when optical flow fails.

In this work, we propose \textbf{SG-RIFE}, a unified architecture that injects dense semantic features from a Vision Foundation Model (DINOv3)~\cite{dinov3} into a flow-based backbone. Our key contributions are:
\begin{itemize}
    \item \textbf{Semantic Injection:} We utilize DINOv3 features to guide the refinement stage. Unlike CLIP~\cite{CLIP} or DINOv2~\cite{dinov2}, DINOv3's Gram Anchoring preserves the spatial consistency required for dense warping.
    \item \textbf{Split-FAPM:} We design a split adapter that compresses features via FiLM~\cite{FiLM} before warping for efficiency and refines them via Squeeze-and-Excitation (SE)~\cite{SE} blocks after warping for artifact suppression.
    \item \textbf{Deformable Semantic Fusion:} We introduce a learnable alignment module that uses deformable convolutions (DCNv2)~\cite{deconv2} to softly align semantic priors with RIFE's pixel-level context, correcting errors near occlusion boundaries.
    \item \textbf{SOTA Efficiency:} SG-RIFE achieves FID~\cite{FID} scores comparable to diffusion-based Consec. BB~\cite{BBD} on SNU-FILM~\cite{snu} while maintaining a runtime of 0.05s per frame.
\end{itemize}

\section{Related Work}
\textbf{Flow-based VFI.} RIFE represents the state-of-the-art in efficiency, utilizing an IFNet for flow estimation and a ContextNet/FusionNet for refinement. While fast, its reliance on local pixel correlations leads to artifacts in disoccluded regions.

\textbf{Generative VFI.} Diffusion models like TLB-VFI~\cite{TLB-VFI} achieve superior perceptual metrics by generating pixels from noise. However, their iterative sampling process incurs high latency (\eg $\sim$22s per frame for LDMVFI~\cite{LDMVFI}), making them unsuitable for real-time tasks.

\textbf{Foundation Models in Low-Level Vision.} Recent works have begun using DINO features for tasks like segmentation~\cite{dinounet} and restoration~\cite{dinoIR}. We extend this to VFI by proposing specific mechanisms (Split-FAPM and DSF) to align these static semantic features with dynamic motion fields.

\section{Method}
\label{sec:method}

\begin{figure*}[t]
  \centering
  \includegraphics[width=0.95\linewidth]{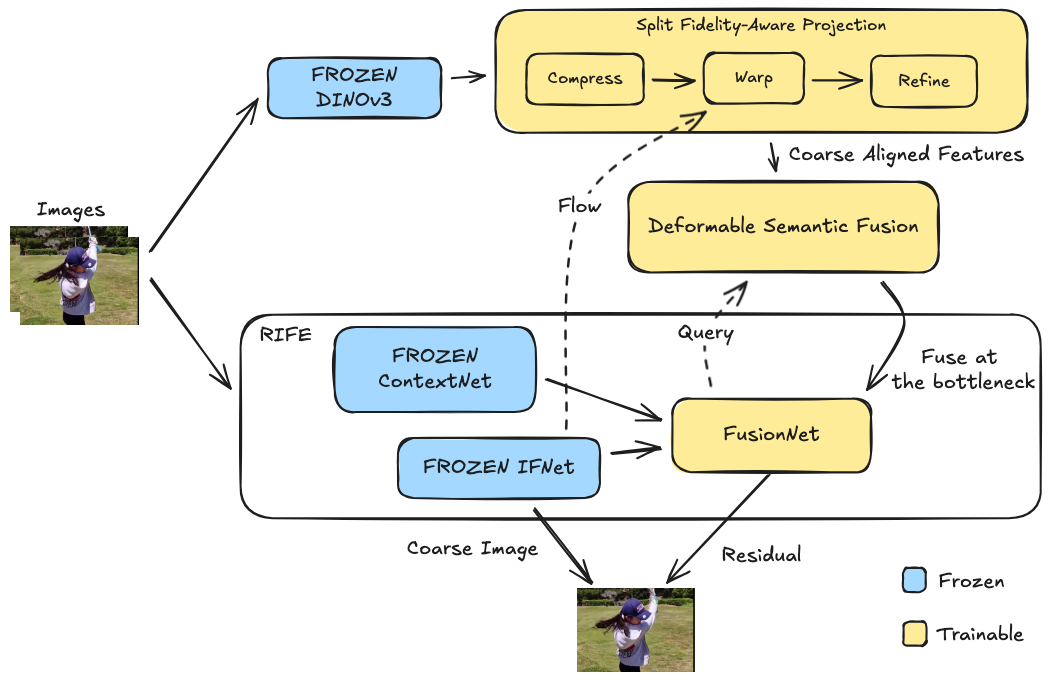} 
  \caption{\textbf{Overview of SG-RIFE.} We extract semantic features from a frozen DINOv3 backbone. These features are compressed via the Split-FAPM, warped using RIFE's intermediate flow, and aligned via Deformable Semantic Fusion (DSF) before being injected into the FusionNet bottleneck.}
  \label{fig:arch}
\end{figure*}

\subsection{The RIFE Backbone}
We build our framework upon RIFE~\cite{huang2020rife} (see Figure~\ref{fig:arch}), which factorizes VFI into two distinct stages: geometry estimation and texture refinement.

\textbf{1. Geometry Estimation (IFNet).}
The Intermediate Flow Network (IFNet) operates in a coarse-to-fine manner. It takes two input frames $I_0, I_1$ and progressively estimates the bidirectional optical flow fields $F_{t\to0}, F_{t\to1}$ at increasing resolutions. Using these flows, it performs backward warping to generate a coarse reconstruction image $\hat{I}_{coarse}$.

\textbf{2. Texture Refinement (ContextNet + FusionNet).}
Since flow-based warping often yields artifacts in occluded regions, RIFE employs a refinement stage. ContextNet extracts pyramidal contextual features from the input frames. FusionNet then takes the coarse reconstruction, the warped context features, and the flow fields to predict a residual map $\Delta$. The final interpolated frame is obtained via:
\begin{equation}
    I_t = \hat{I}_{coarse} + \Delta
\end{equation}
In our SG-RIFE framework, we \textbf{freeze IFNet and ContextNet} to preserve the pre-trained motion and texture priors. We efficiently inject semantic guidance by fine-tuning only the FusionNet and our lightweight adapters, allowing the model to inpaint plausible textures in the residual $\Delta$ where the coarse flow failed.

\subsection{Semantic Feature Extraction}
\begin{figure}[t]
  \centering
  \includegraphics[width=\linewidth]{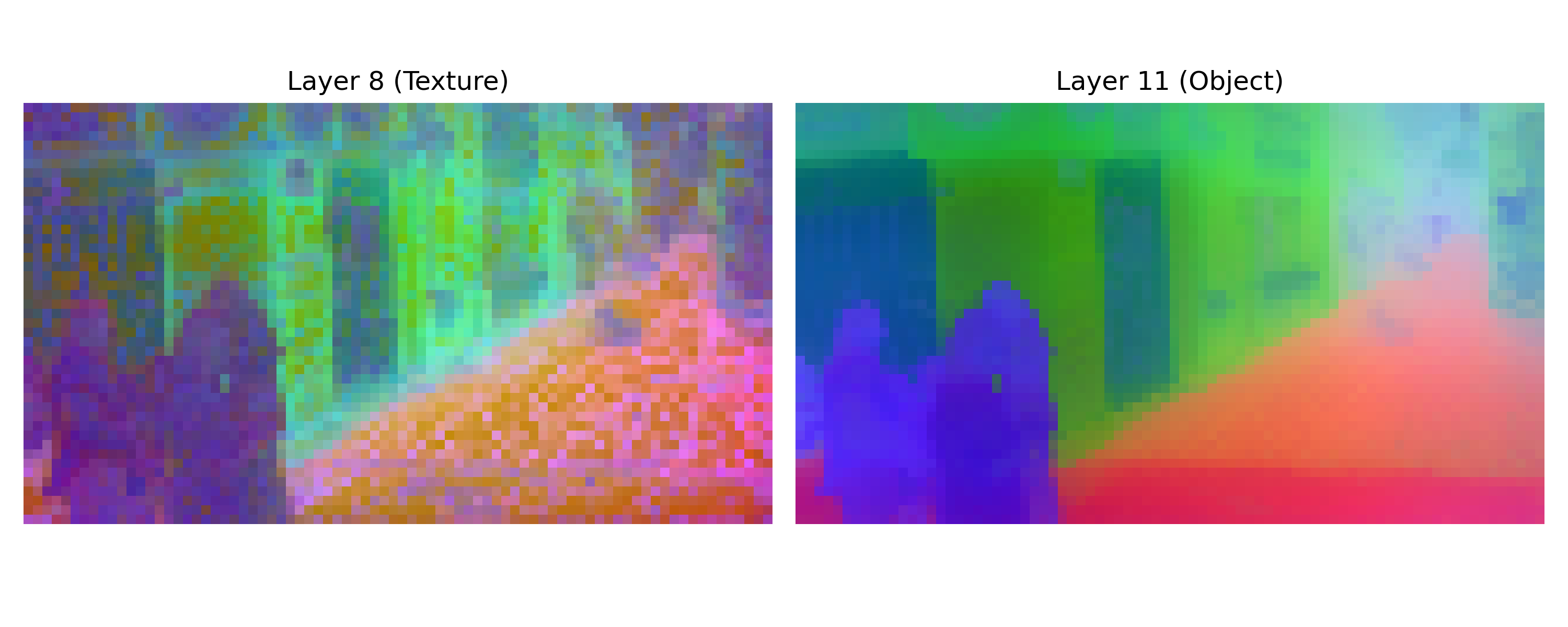}
  \caption{\textbf{Hierarchical Feature Selection.} Visualization of principal components from DINOv3 layers. Layer 8 (Left) exhibits high-frequency variance corresponding to local textural patterns and edges. Layer 11 (Right) demonstrates semantic stability, providing semantically coherent guidance that helps the FusionNet preserve global object boundaries.}
  \label{fig:layers}
\end{figure}
We employ DINOv3~\cite{dinov3} as our semantic backbone. We select DINOv3 over DINOv2~\cite{dinov2}/CLIP~\cite{CLIP} because its Gram Anchoring mechanism explicitly regularizes spatial consistency, preventing the localization loss typical in large ViTs. We extract hidden states from intermediate layers (indices 8 and 11 for DINOv3-Small) rather than the output. This strategy, inspired by DINO-IR~\cite{dinoIR}, allows us to capture both local texture details and global object semantics (Figure~\ref{fig:layers}), which is critical for restoration tasks.

\subsection{Split Fidelity-Aware Projection (Split-FAPM)}
Directly processing high-dimensional ViT features ($C=384$) is computationally prohibitive for dense warping. Drawing inspiration from the Fidelity-Aware Projection Module (FAPM) in Dino U-Net~\cite{dinounet}, we introduce a lightweight Split-FAPM. We split the module into two stages to optimize for the intermediate flow warping step.

\textbf{1. Pre-Warp Compression (Domain Adaptation).} 
To adapt the frozen DINO features to the VFI domain, we employ the dual-branch decomposition strategy from Dino U-Net~\cite{dinounet}. The input $D_{raw}$ is fed into two parallel $1\times1$ convolutional branches:
\begin{itemize}
    \item \textbf{Feature Branch (Specific):} Extracts the latent feature map $F_{feat}$ using task-specific bases.
    \item \textbf{Modulation Branch (Shared):} Uses a shared basis to predict affine parameters $\gamma$ and $\beta$ for Feature-wise Linear Modulation (FiLM)~\cite{FiLM}.
\end{itemize}
The compressed features are obtained via:
\begin{equation}
    F_{compressed} = \gamma \odot F_{feat} + \beta
\end{equation}
This projects the semantic features into the task-specific manifold while reducing the channel dimension ($384 \to 256$), ensuring the features match the FusionNet bottleneck capacity while lowering the memory cost of the subsequent warping.

\textbf{2. Post-Warp Refinement (Artifact Suppression).} 
\label{sec:post-warp}
Warping latent features often introduces spatial discontinuities. After the warping stage (Sec.~\ref{sec:alignment}), we apply a residual refinement block. This block serves two purposes:
\begin{itemize}
    \item \textbf{Channel Matching:} It projects the warped features (256-dim) to the specific channel dimensions required by the multi-scale FusionNet layers (256 for $S_3$, 128 for $S_2$).
    \item \textbf{Feature Restoration:} It employs Depthwise Separable Convolutions to repair spatial breaks and Squeeze-and-Excitation (SE) blocks~\cite{SE} for channel-wise recalibration. Following RIFE~\cite{huang2020rife}, we remove Batch Normalization to avoid distorting motion statistics, and we replace ReLU with GELU to ensure activation compatibility with the DINOv3 backbone.
\end{itemize}

\subsection{Flow-Guided Deformable Alignment}
\label{sec:alignment}
\begin{figure*}[t]
  \centering
  \includegraphics[width=\linewidth]{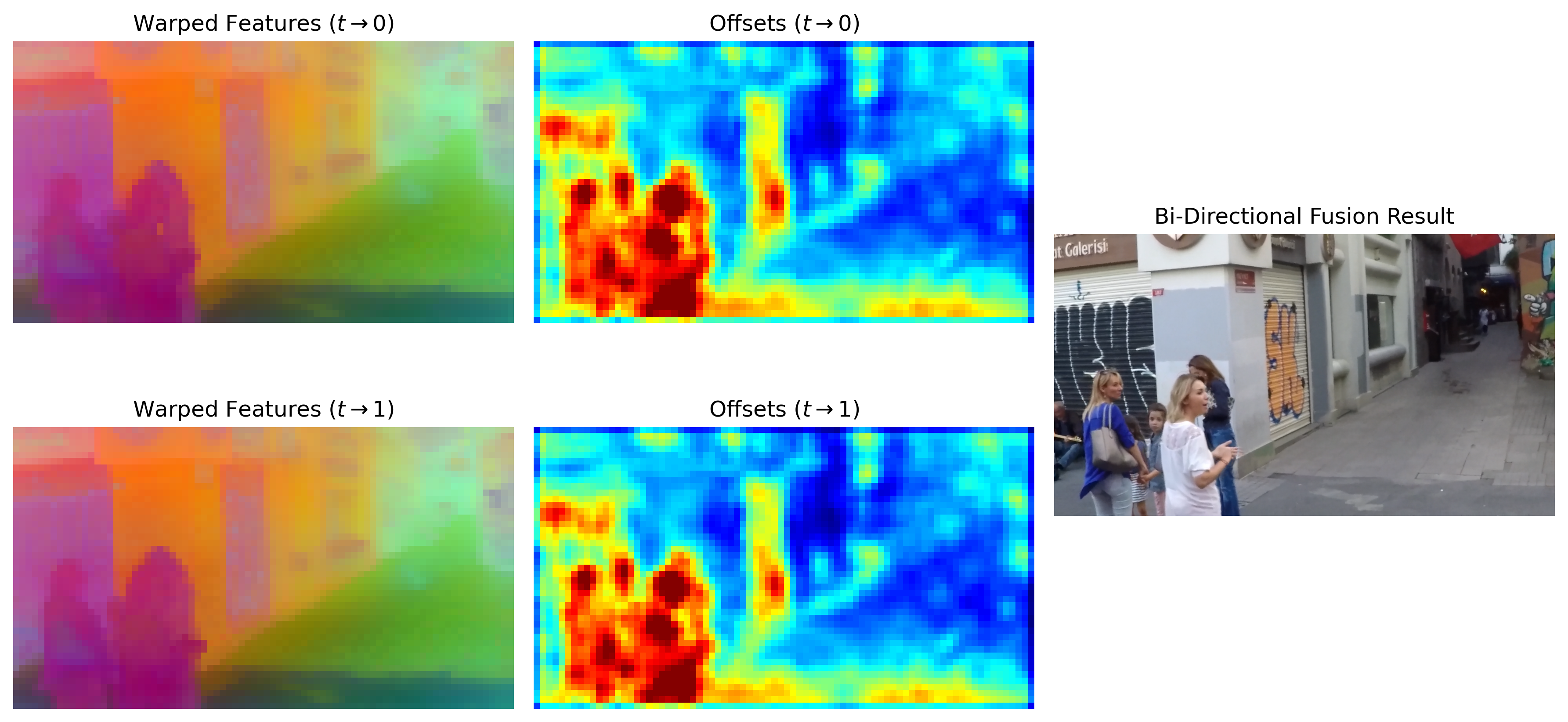}
  \caption{\textbf{Visualization of the Flow-Guided Deformable Alignment (DSF) Mechanism.} \textit{Left:} Coarsely warped semantic features exhibit misalignment due to optical flow errors. \textit{Center:} The offset magnitude maps ($||\Delta p||$) reveal the active alignment regions. Note the high activation (red) on the dynamic foreground subjects, indicating that the module is performing correction to compensate for flow inaccuracies. \textit{Right:} The final bi-directional fusion result demonstrates seamless integration of the corrected features.}
  \label{fig:mechanism}
\end{figure*}
Rigid warping is often inaccurate near occlusion boundaries (Figure~\ref{fig:mechanism}) and leads to misalignment between semantic and pixel features. We introduce a \textbf{Deformable Semantic Fusion (DSF)} module that uses modulated deformable convolutions (DCNv2)~\cite{deconv2} to softly align the refined DINO features to RIFE's context.

First, we obtain coarse semantic priors by warping the compressed features and passing them through the Split-FAPM Refiner (Sec. \ref{sec:post-warp}) to correct spatial discontinuities:
\begin{equation}
\begin{split}
    \hat{D}_{t \leftarrow 0} &= \Phi_{ref}(\mathcal{W}(D_0, F_{t \to 0})),\\
    \hat{D}_{t \leftarrow 1} &= \Phi_{ref}(\mathcal{W}(D_1, F_{t \to 1}))
\end{split}
\end{equation}
where $\Phi_{ref}$ denotes the residual refinement block. To predict interaction-aware offsets, we then project RIFE's pixel context $F_{ctx}$ and these refined features $\hat{D}$ into a shared Query-Key-Value space:
\begin{equation}
    Q = \phi_q(F_{ctx}),\quad K = \phi_k(\hat{D}),\quad V = \phi_v(\hat{D}),
\end{equation}
where $\phi_k, \phi_v$ are shared weights across past and future branches. Offsets $\Delta p$ and modulation scalars $\Delta m$ are predicted by 
a convolution over the concatenated $[Q, K]$. The aligned features 
are synthesized via DCNv2 and gated by a learnable scalar $\gamma$:
\begin{equation}
    F_{aligned}^{(g)}(p) = \gamma \cdot \sum_{k=1}^{9} \mathbf{W}_k^{(g)} \cdot V^{(g)}(p + p_k + \Delta p_k^{(g)}) \cdot \Delta m_k^{(g)}
\end{equation}
where $g$ indexes the deformable groups with kernel size $3\times3$. 
Following EDVR~\cite{edvr}, we use grouped deformable convolutions 
for computational efficiency, reducing parameters from $C^2 \times 9$ 
to $C^2/G \times 9$ where $G$ is the number of groups. We deploy 
two DSF modules in the FusionNet: one at the intermediate encoder 
level ($S_2$) with 4 groups processing 128-channel features, and 
one at the bottleneck ($S_3$) with 8 groups processing 256-channel 
features, maintaining 32 channels per group across scales. We 
initialize $\gamma=0$, allowing the model to progressively admit 
semantic gradients only as the alignment stabilizes.

\subsection{Semantic-Aware Refinement}
Aligned semantic features are injected into the FusionNet to guide texture inpainting. We use a selective multi-scale injection: shallow DINO features (layer 8) are fused at an intermediate encoder level to guide local textures, while deeper features (layer 11) are injected at the bottleneck to provide global object context. We do not upsample and inject semantics into the highest-resolution layers; instead, RIFE's ContextNet handles fine edges directly from RGB inputs, avoiding noisy upsampled semantics and extra cost.

\subsection{Objective Function}
\label{sec:objective}

Our training objective combines the losses from RIFE with two regularization terms. The total loss $\mathcal{L}_{total}$ is defined as:
\begin{equation}
\begin{split}
    \mathcal{L}_{total} = & \lambda_{rec}\mathcal{L}_{rec} + \lambda_{dis}\mathcal{L}_{dis} + \lambda_{tea}\mathcal{L}_{tea}\\
     & + \lambda_{sem}\mathcal{L}_{sem} + \lambda_{reg}\mathcal{L}_{reg}
\end{split}
\end{equation}
where $\mathcal{L}_{rec}$ and $\mathcal{L}_{tea}$ are the Laplacian pyramid reconstruction losses ~\cite{huang2020rife}, and $\mathcal{L}_{dis}$ is the student-teacher distillation loss. We introduce:
\begin{itemize}
    \item \textbf{Semantic Consistency Loss ($\mathcal{L}_{sem}$):} An $L_1$ penalty ensuring the interpolated frame's semantic features match the ground truth: $\| D(I_{pred}) - D(I_{gt}) \|_1$.
    \item\textbf{Offset Regularization ($\mathcal{L}_{reg}$):} To prevent divergent warping in the DSF modules, we regularize the predicted offsets from both injection points:  $\| \Delta p_{S_2} \|_1 + \| \Delta p_{S_3} \|_1$~\cite{understandingDA}.
\end{itemize}
Empirically, we set the weights to $\lambda_{rec}=0.1$, $\lambda_{dis}=0.01$, $\lambda_{tea}=0.1$, $\lambda_{sem}=0.5$, and $\lambda_{reg}=0.0001$, prioritizing semantic fidelity ($\lambda_{sem}$) over pure pixel reconstruction.
\begin{table*}[t]
\centering
\caption{\textbf{Quantitative Evaluation on SNU-FILM.} Best results are \textbf{bold}, second-best are \underline{underlined}. Metrics are reported as PSNR$\uparrow$/SSIM$\uparrow$ and LPIPS$\downarrow$/FloLPIPS$\downarrow$/FID$\downarrow$. $^{\dagger}$ indicates results adopted from the TLB-VFI~\cite{TLB-VFI} benchmark. Runtime measures seconds per frame to interpolate a 480 × 720 image.  RIFE and SG-RIFE runtimes are measured on an L4 GPU; others are sourced from original reports (A5000).}
\resizebox{\textwidth}{!}{
\setlength{\tabcolsep}{3.5pt} 
\begin{tabular}{l cc cc cc cc c}
\toprule
& \multicolumn{2}{c}{\textbf{SNU-FILM (Easy)}} 
& \multicolumn{2}{c}{\textbf{SNU-FILM (Medium)}} 
& \multicolumn{2}{c}{\textbf{SNU-FILM (Hard)}} 
& \multicolumn{2}{c}{\textbf{SNU-FILM (Extreme)}} 
& \multirow{2}{*}{Time (s)} \\
\cmidrule(r){2-3} \cmidrule(lr){4-5} \cmidrule(lr){6-7} \cmidrule(l){8-9}
Method & PSNR/SSIM$\uparrow$ & LPIPS/FloLPIPS/FID$\downarrow$ & PSNR/SSIM$\uparrow$ & LPIPS/FloLPIPS/FID$\downarrow$ & PSNR/SSIM$\uparrow$ & LPIPS/FloLPIPS/FID$\downarrow$ & PSNR/SSIM$\uparrow$ & LPIPS/FloLPIPS/FID$\downarrow$ & \\
\midrule
\multicolumn{10}{l}{\textit{Regression-based}} \\
RIFE'22~\cite{huang2020rife} & 40.018 / \textbf{0.991} & 0.018 / 0.030 / 5.456 & 35.719 / 0.979 & 0.031 / 0.059 / 10.833 & 30.075 / 0.933 & 0.066 / 0.124 / 23.320 & 24.819 / 0.853 & 0.139 / 0.232 / 47.458 & \textbf{0.01} \\
VFIformer'22~\cite{VFIformer}$^{\dagger}$ & \underline{40.130} / \textbf{0.991} & 0.018 / 0.029 / 5.918 & 36.090 / \underline{0.980} & 0.033 / 0.053 / 11.271 & 30.670 / 0.938 & 0.061 / 0.100 / 22.775 & 25.430 / 0.864 & 0.119 / 0.185 / 40.586 & 4.34 \\
IFRNet'22~\cite{IFRNet}$^{\dagger}$ & 40.100 / \textbf{0.991} & 0.021 / 0.031 / 6.863 & \underline{36.120} / \underline{0.980} & 0.034 / 0.050 / 12.197 & 30.630 / 0.937 & 0.059 / 0.093 / 23.254 & 25.270 / 0.861 & 0.116 / 0.182 / 42.824 & 0.10 \\
AMT'23~\cite{AMT}$^{\dagger}$ & 39.880 / \textbf{0.991} & 0.022 / 0.034 / 6.139 & \underline{36.120} / \textbf{0.981} & 0.035 / 0.055 / 11.039 & 30.780 / \textbf{0.939} & 0.060 / 0.092 / 20.810 & 25.430 / \underline{0.865} & 0.112 / 0.177 / 40.075 & 0.11 \\
UPR-Net'23~\cite{UPRNet}$^{\dagger}$ & \textbf{40.440} / \textbf{0.991} & 0.018 / 0.029 / 5.669 & \textbf{36.290} / \underline{0.980} & 0.034 / 0.052 / 10.983 & \underline{30.860} / 0.938 & 0.062 / 0.097 / 22.127 & \underline{25.630} / 0.864 & 0.112 / 0.176 / 40.098 & 0.70 \\
EMA-VFI'23~\cite{EMAVFI}$^{\dagger}$ & 39.980 / \textbf{0.991} & 0.019 / 0.038 / 5.882 & 36.090 / \underline{0.980} & 0.033 / 0.053 / 11.051 & \textbf{30.940} / \textbf{0.939} & 0.060 / 0.091 / 20.679 & \textbf{25.690} / \textbf{0.866} & 0.114 / \underline{0.170} / 39.051 & 0.72 \\
\midrule
\multicolumn{10}{l}{\textit{Generative-based}} \\
MCVD'22~\cite{MCVD}$^{\dagger}$ & 22.201 / 0.828 & 0.199 / 0.230 / 32.246 & 21.488 / 0.812 & 0.213 / 0.243 / 37.474 & 20.314 / 0.766 & 0.250 / 0.292 / 51.529 & 18.464 / 0.694 & 0.320 / 0.385 / 83.156 & 52.55 \\
LDMVFI'24~\cite{LDMVFI}$^{\dagger}$ & 38.890 / 0.988 & 0.014 / 0.024 / 5.752 & 33.975 / 0.971 & 0.028 / 0.053 / 12.485 & 29.144 / 0.911 & 0.060 / 0.114 / 26.520 & 23.349 / 0.827 & 0.123 / 0.204 / 47.042 & 22.32 \\
PerVFI'24~\cite{pervfi}$^{\dagger}$ & 38.065 / 0.986 & 0.014 / 0.022 / 5.917 & 34.588 / 0.973 & 0.024 / 0.040 / 10.395 & 29.821 / 0.928 & \underline{0.046} / \textbf{0.077} / 18.887 & 25.033 / 0.854 & \textbf{0.090} / \textbf{0.151} / \underline{32.372} & 1.52 \\
Consec. BB'24~\cite{BBD}$^{\dagger}$ & 39.637 / 0.990 & \textbf{0.012} / \underline{0.019} / 4.791 & 34.886 / 0.974 & \underline{0.022} / \underline{0.039} / 9.039 & 29.615 / 0.929 & 0.047 / 0.091 / 18.589 & 24.376 / 0.848 & 0.104 / 0.184 / 36.631 & 2.60 \\
TLB-VFI'25~\cite{TLB-VFI}$^{\dagger}$ & 39.460 / 0.990 & \textbf{0.012} / \textbf{0.018} / \underline{4.658} & 35.308 / 0.977 & \textbf{0.021} / \textbf{0.036} / \textbf{8.518} & 29.529 / 0.929 & \textbf{0.044} / \underline{0.085} / \textbf{17.470} & 24.513 / 0.847 & \underline{0.095} / \textbf{0.151} / \textbf{29.868} & 0.69 \\
\midrule
\multicolumn{10}{l}{\textit{Proposed}} \\
\textbf{SG-RIFE (Ours)} & 39.946 / 0.990 & \textbf{0.012} / \underline{0.019} / \textbf{4.557} & 35.495 / 0.977 & \textbf{0.021} / 0.040 / \underline{8.678} & 29.736 / 0.929 & 0.047 / 0.090 / \underline{17.896} & 24.439 / 0.844 & 0.107 / 0.184 / 36.272 & \underline{0.05} \\
\bottomrule
\end{tabular}}
\label{tab:snu_comparison}
\end{table*}
\section{Experiments}
\label{sec:experiments}

\noindent\textbf{Datasets and Evaluation.}
We train SG-RIFE on the Vimeo90K~\cite{vimeo90k} training split, which consists of 51,312 triplets with fixed resolution $448 \times 256$. We use standard augmentations including random cropping, horizontal/vertical flipping, and temporal reversal.
For evaluation, we report results on the SNU-FILM~\cite{snu} benchmark, which contains four difficulty splits (Easy, Medium, Hard, Extreme) based on motion magnitude. To evaluate perceptual realism and temporal coherence against diffusion baselines, we report LPIPS~\cite{lpips}, FloLPIPS~\cite{FLOLPIPS}, and FID~\cite{FID} .

\vspace{1mm}
\noindent\textbf{Implementation Details.} 
Our model is implemented in PyTorch. We initialize the semantic backbone with DINOv3-Small pretrained on LVD-1689M and the flow backbone with official RIFE checkpoints. The adapters and fusion layers are initialized using He initialization~\cite{HeInit}. 

We adopt a two-stage training strategy to ensure stable convergence:
\begin{enumerate}
    \item \textbf{Stage 1 (Alignment):} We train only the Split-FAPM adapters and DSF module for 5 epochs. This forces the model to learn proper feature alignment without perturbing the fusion weights.
    \item \textbf{Stage 2 (Fusion):} We unfreeze the FusionNet and train jointly with the adapters for 25 epochs (total 30 epochs) to learn optimal semantic injection.
\end{enumerate}
We use the AdamW optimizer~\cite{adamw} with a batch size of 64, a learning rate of $2 \times 10^{-4}$, and cosine annealing. To further ensure stability during the sensitive warping updates, we apply gradient clipping with a maximum norm of 1.0. The entire training process is conducted on a single NVIDIA L4 GPU.

\subsection{Quantitative Analysis}
We evaluate SG-RIFE using a combination of fidelity and perceptual metrics. While PSNR and SSIM measure pixel-level reconstruction accuracy, they favor "safe" blurring over high-frequency details in complex motion. To capture the visual realism required for high-quality VFI, we utilize FID to assess global distribution similarity and LPIPS to measure local perceptual features. Furthermore, we report FloLPIPS to specifically evaluate temporal consistency and the model's ability to maintain coherent structures during flow-guided warping.

Table~\ref{tab:snu_comparison} presents a comprehensive evaluation of SG-RIFE across the four difficulty tiers of the SNU-FILM benchmark. The results demonstrate that our semantic injection strategy effectively narrows the perceptual gap between deterministic regression models and heavy generative baselines.

\noindent\textbf{Perceptual Performance.} SG-RIFE demonstrates a notable improvement in perceptual metrics compared to the RIFE backbone and other real-time models. On the SNU-Easy split, our method achieves an FID of 4.557, outperforming recent diffusion-based models such as Consec.~BB (4.791) and TLB-VFI (4.658). As motion complexity increases (SNU-Hard), SG-RIFE maintains a significant lead over real-time baselines and surpasses LDMVFI in FID by a margin of 8.62 (17.896 vs. 26.520). Furthermore, SG-RIFE maintains competitive FloLPIPS scores compared to both the RIFE baseline and diffusion methods, suggesting that the injection of high-level semantic priors does not introduce additional temporal instability, ensuring that synthesized textures remain coherent over time.

\noindent\textbf{Perception-Distortion Trade-off.} In accordance with the perception-distortion trade-off~\cite{blau2018perception}, the prioritization of semantic sharpness results in a marginal decrease in fidelity metrics compared to the RIFE backbone (\eg a PSNR drop of 0.339~dB on the Hard split). We observe that the weighting of the semantic consistency loss ($\lambda_{sem}$) serves as a critical control for this trade-off: lowering $\lambda_{sem}$ shifts the model toward a conservative regression-based behavior, yielding higher PSNR and SSIM at the cost of diminished perceptual "sharpness" (higher FID/LPIPS). By maintaining a higher $\lambda_{sem}$, SG-RIFE utilizes DINOv3 priors to synthesize high-frequency textures that are perceptually superior, as validated by our competitive FID scores.

\noindent\textbf{Efficiency and Complexity.} A primary advantage of SG-RIFE is its parameter-efficient design. As detailed in Table~\ref{tab:param_breakdown}, by freezing the computation-heavy semantic (DINOv3) and motion (IFNet) backbones, the trainable components consist of only 5.6M parameters ($\approx 16\%$ of total inference parameters). This allows SG-RIFE to maintain a runtime of 0.05s per frame on a single L4 GPU. Compared to diffusion-based Consec. BB (2.6s) and TLB-VFI (0.69s), SG-RIFE is approximately 52$\times$ and 14$\times$ faster, respectively, while achieving comparable perceptual fidelity.

\begin{table}[t]
    \centering
    \small 
    \caption{\textbf{Parameter Efficiency Analysis.} We freeze the heavy feature extractors (DINOv3 and IFNet) and only train the lightweight adapters and fusion modules. Note that we achieve perceptual quality competitive with recent diffusion-based VFI models on standard benchmarks by updating only \textbf{16\%} of the total parameters.}
    \label{tab:param_breakdown}
    \begin{tabular}{lcr}
        \toprule
        \textbf{Module} & \textbf{Status} & \textbf{Params (M)} \\
        \midrule
        DINOv3 (Small) & Frozen & 21.6 \\
        IFBlocks (0--2) & Frozen & 7.5 \\
        ContextNet & Frozen & 0.3 \\
        \midrule
        \textit{Total Frozen} & & \textit{29.4} \\
        \midrule
        Split-FAPM (Compressor) & Trained & 0.5 \\
        Split-FAPM (Refiner) & Trained & 0.3 \\
        FusionNet (Base) & Fine-tuned$^{\dagger}$ & 2.0 \\
        FusionNet (Injection) & Trained & 2.8 \\
        \midrule
        \textit{Total Trainable} & & \textbf{5.6} \\
        \midrule
        \textbf{Total Inference} & & \textbf{34.4$^{\star}$} \\
        \bottomrule
        \multicolumn{3}{l}{\scriptsize $^{\dagger}$Frozen during Stage 1 warmup.} \\
        \multicolumn{3}{l}{\scriptsize $^{\star}$Excludes Teacher IFBlock (training only).} \\
    \end{tabular}
\end{table}

\subsection{Qualitative Analysis}
\begin{figure}[t]
    \centering
    \setlength{\tabcolsep}{0.5pt}
    
    \vspace{2mm} 
    
    \begin{tabular}{cccc}
        \includegraphics[width=0.245\linewidth]{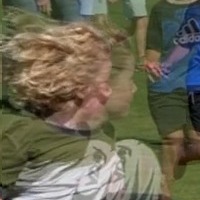} &
        \includegraphics[width=0.245\linewidth]{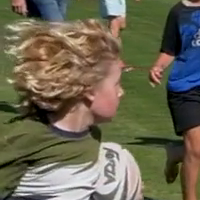} &
        \includegraphics[width=0.245\linewidth]{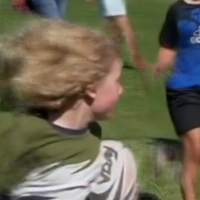} &
        \includegraphics[width=0.245\linewidth]{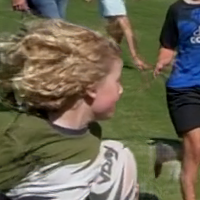} \\
        
        \includegraphics[width=0.245\linewidth]{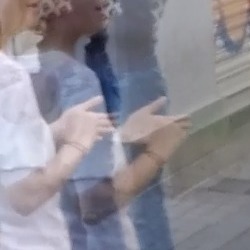} &
        \includegraphics[width=0.245\linewidth]{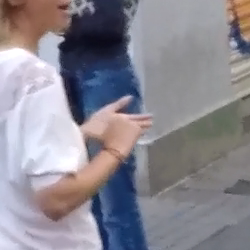} &
        \includegraphics[width=0.245\linewidth]{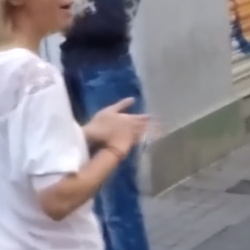} &
        \includegraphics[width=0.245\linewidth]{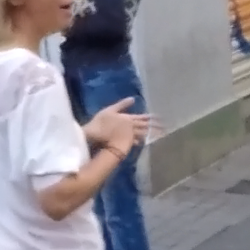} \\
        
        \scriptsize (a) Overlaid Inputs & \scriptsize (b) Ground Truth & \scriptsize (c) RIFE (Base) & \scriptsize (d) \textbf{SG-RIFE (Ours)} \\
    \end{tabular}
    
    \caption{\textbf{Qualitative Comparison on the SNU-FILM (Extreme) dataset.} 
    (a) The overlaid input frames ($I_0$ and $I_1$) illustrate the large motion. 
    (c) The baseline RIFE suffers from ghosting. 
    (d) Our SG-RIFE successfully refines high-frequency textures and mitigates ghosting.}
    \label{fig:qualitative_snu}
\end{figure}

Qualitative results are shown in Figure~\ref{fig:qualitative_snu}. The overlaid inputs (a) reveal large motion magnitudes. As seen in (c), RIFE fails to track the fast-moving object, resulting in blurred regions (\eg hair and hands). In contrast, SG-RIFE (d) leverages the DSF to inpaint plausible textures. However, since SG-RIFE operates as a semantic residual refinement module, it is constrained by the coarse flow estimation of the frozen IFNet.

\subsection{Limitations and Future Directions}
\noindent\textbf{Experimental Scope.} Due to computational resources, we focus our evaluation on the SNU-FILM benchmark to demonstrate the efficacy of semantic injection. Future work should extend validation to additional benchmarks (\eg UCF101, Xiph) to assess generalization across broader motion patterns.

\noindent\textbf{Ablation Studies.} While we demonstrate that the complete system achieves competitive results, comprehensive ablations of individual components (Split-FAPM contribution, DSF effectiveness, semantic backbone comparison) would strengthen understanding of each module's contribution and remain for future investigation.

\noindent\textbf{Flow Backbone Dependency.} Our approach relies on the frozen IFNet for motion estimation. In scenarios of catastrophic flow failure (\eg extreme non-linear motion or severe occlusion), semantic guidance cannot correct fundamental trajectory errors. As observed in Figure~\ref{fig:qualitative_snu}, this can result in sharp but spatially misaligned details.

\section{Conclusion}
We presented SG-RIFE, a semantic-guided VFI architecture. By augmenting a flow-based backbone with DINOv3 features via Split-FAPM and DSF, we achieve perceptual quality competitive with recent diffusion baselines while maintaining near real-time performance. This work demonstrates that explicit semantic injection offers a highly efficient alternative to generative diffusion for video enhancement, proving that high-level object priors can effectively mitigate the limitations of traditional flow-based warping in complex motion scenarios.

{\small
\bibliographystyle{ieee_fullname}
\bibliography{egbib}
}

\end{document}